\documentclass[manuscript]{ieeeconf}  

\IEEEoverridecommandlockouts                              

\usepackage{tikz}
\usetikzlibrary{shapes.geometric,arrows,calc}

\title{\LARGE \bf
Bot Appétit! Exploring how Robot Morphology Shapes Perceived Affordances via a Mise en Place Scenario in a VR Kitchen
}

\author{Rachel Ringe\orcidlink{0009-0005-4696-5873}\textsuperscript{1*}, Leandra Thiele\orcidlink{0000-0002-8074-8507}\textsuperscript{2}, Mihai Pomarlan\orcidlink{0000-0002-1304-581X}\textsuperscript{2}, Nima Zargham\orcidlink{0000-0003-4116-0601}\textsuperscript{1}, Robin Nolte\orcidlink{0009-0004-2975-6378}\textsuperscript{1}, \\Lars Hurrelbrink\orcidlink{0009-0007-4316-4047}\textsuperscript{1}, Rainer Malaka\orcidlink{0000-0001-6463-4828}\textsuperscript{1}
\thanks{This work was funded by the FET-Open Project \#951846 ``MUHAI – Meaning and Understanding for Human-centric AI'' by the EU Pathfinder and Horizon 2020 Program.
The research reported in this paper has been (partially) supported by the German Research Foundation DFG, as part of Collaborative Research Center (Sonderforschungsbereich) 1320 Project-ID 329551904 ``EASE  ---  Everyday Activity Science and Engineering'', University of Bremen (\protect\url{https://www.ease-crc.org}).
The research was conducted in subprojects ``P05-N  ---  Principles of Metareasoning for Everyday Activities'' and ``P01 --- Embodied semantics for the language of action and change: Combining analysis, reasoning and simulation''.
}
\thanks{*Corresponding author: {\tt\small \href{mailto:rringe@uni-bremen.de}{rringe@uni-bremen.de}}}
\thanks{$^{1}$Digital Media Lab, University of Bremen, Germany}%
\thanks{$^{2}$Department of Linguistics, University of Bremen, Germany}%
}

\let\labelindent\relax
\usepackage{pgfplots}
\pgfplotsset{compat=1.17}
\usetikzlibrary{patterns}
\usetikzlibrary{backgrounds}
\definecolor{cbgreen}{HTML}{1B9E77}
\definecolor{cborange}{HTML}{D95F02}
\definecolor{cbpurple}{HTML}{7570B3}
\usepackage{tabularx} 
\usepackage{wrapfig} 
\usepackage{graphicx} 
\usepackage{url} 
\usepackage{hyperref} 
\usepackage{cleveref}
\usepackage[]{footmisc} 
\usepackage{orcidlink}
\usepackage[nopatch=eqnum]{microtype}
\usepackage{enumitem}
\usepackage{comment}
\usepackage{caption}
\usepackage{subcaption}
\usepackage{microtype}

\definecolor{verylightgray}{gray}{0.90}

\tikzstyle{process} = [minimum width = 4.1cm, line width=1pt, minimum height = 0.6cm, rectangle, rounded corners, text centered, draw=black, fill=verylightgray, font=\scriptsize\sffamily]
\tikzstyle{arrow} = [thick,->,>=stealth, line width=1pt]

\begin{document}

\maketitle

\thispagestyle{empty}
\pagestyle{empty}


\begin{abstract}
%
This study explores which factors of the visual design of a robot may influence how humans would place it in a collaborative cooking scenario and how these features may influence task delegation. Human participants were placed in a Virtual Reality (VR) environment and asked to set up a kitchen for cooking alongside a robot companion while considering the robot's morphology. We collected multimodal data for the arrangements created by the participants, transcripts of their think-aloud as they were performing the task, and transcripts of their answers to structured post-task questionnaires. Based on analyzing this data, we formulate several hypotheses: humans prefer to collaborate with biomorphic robots; human beliefs about the sensory capabilities of robots are less influenced by the morphology of the robot than beliefs about action capabilities; and humans will implement fewer avoidance strategies when sharing space with gracile robots. We intend to verify these hypotheses in follow-up studies. 


\end{abstract}

\section{INTRODUCTION}

Cooking involves multiple tools and subtasks of varying complexity depending on the recipe and is often performed collaboratively. As simple household robots like vacuum cleaners gain popularity, humans will likely share their kitchens with robotic helpers in the future \cite{zhu2020robot,Zhi2025}. While substantial research focuses on functionality -- enabling robots to fetch and pour ingredients -- research should also explore human reactions to the presence of a robot in their kitchen and preferences regarding their features and collaboration patterns.
Robot morphology significantly influences people's first impressions;
e.g., anthropomorphic appearance can increase user trust \cite{natarajan2020effects}.
These effects extend beyond coarse categories~\cite{abot}; even isolated morphological features -- sub-components such as manipulators and surfaces -- impact perceived task suitability \cite{tenhundfeld2020robot}.

%
Recent efforts, like the \textit{MetaMorph} model \cite{MetaMorph}, have begun standardizing coding of morphological features beyond the discretion of study designers.
Nevertheless, practical limitations persist -- real robots are costly, and modifying even single features is complex, constraining studies either to study more readily modifiable visual aspects such as behavior (e.g.,~\cite{Shilon2024}) or rely on databases of static images (e.g., \cite{abot,tenhundfeld2020robot,HWANG2013459}).
Furthermore, precise predictions about the perceptual effects remain challenging.
For instance, it is not obvious that a robot combining a cylindrical head with a human-like torso and limbs would necessarily appear most extroverted compared to similar silhouettes, but with different component shapes~\cite{HWANG2013459}.
The hypothesis space is virtually unrestricted.
A massive robot might feel threatening, a frail one incapable. 

\begin{figure}[tb]
    \centering
    \includegraphics[width=\columnwidth]{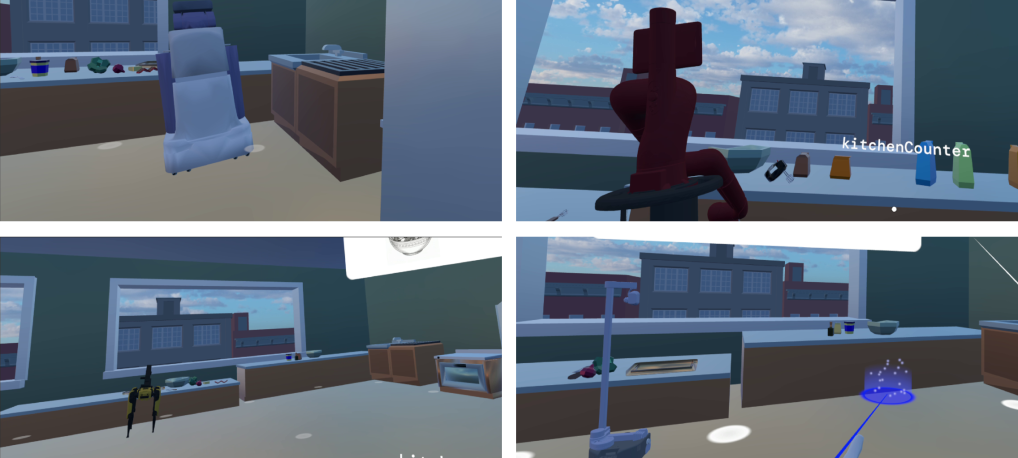}
    \caption{Example arrangements produced by human participants in a VR kitchen. Some arrangements show separations into work stations, with robot and human kept wide apart (top two examples), whereas others mix task-related items and would involve closer human-robot collaboration (bottom).}
    \label{fig:arrangements}
\end{figure}

In summary, understanding how visual design choices impact HRI is critical for working towards successful collaboration.
Thus, we raise the following research questions: which features of a robot's appearance have a significant impact on a human's readiness to trust a robot with a task? Which features have a significant impact on a human's readiness to trust the robot enough to collaborate on a task in shared proximity? In this paper, we address them by exploratively investigating how robot morphology influences user expectations, specifically in collaborative contexts. 
We employ a virtual Reality (VR) kitchen environment, as cooking tasks are popular in Human-Robot Interaction (HRI) for their complexity.
Furthermore, it might be the only everyday scenario in which a robot would routinely handle knives -- a tool that can double as a weapon -- rendering it a uniquely rich setting for examining user concerns, task delegation, and safety perceptions.
VR enables systematic manipulation of robot morphology, such as varying robot size or arm count, without physical constraints, and also allows for easy tracking of object placement via their coordinates. It also allows using a larger variety of robot platforms than would be otherwise financially feasible, and to do so in a setting that is safe for the user.
VR-simulated field studies are a powerful research tool where largely similar behavior to real settings have been observed~\cite{Ville2020,Paneva2020,ZarghamMulti}. Researchers recommend designing scenarios that foster natural behavior and user-driven exploration~\cite{Ville2020}.


Participants were presented with a \textit{mise en place} scenario -- the preparatory phase preceding cooking -- requiring them to distribute tasks and position items, including determining optimal placements for the robot itself.
%
%
%
Our contributions are as follows:
\begin{itemize}
    \item We have collected a multimodal dataset (transcripts of think-alouds, VR video recordings, object 3D coordinates) from 22 participants tasked with preparing a kitchen for collaboration with one of several robots. This dataset is openly available \footnote{\label{Suppl.mat.} \url{https://github.com/ease-crc/BotAppetit}}.
    \item We qualitatively assessed how individual morphological features shape perceived robot affordances and collaborative potential by analyzing the think-aloud-protocols (TAPs), with a view towards identifying which features are perceived as most relevant.
    \item We quantitatively analyzed the task assignments the participants provided, to uncover possible tendencies in how robot appearance influences user readiness to collaborate with a robot.
\end{itemize}

%
 
\section{RELATED WORK}
\label{sec:related_work}
\subsection{Robot Appearance and User Perception}
Appearance is one of the first aspects humans notice when encountering robots \cite{Haring2016How}, shaping initial impressions and influencing subsequent interactions. Indeed, extensive research has shown that a robot's visual design significantly impacts Human-Robot Interaction \cite{riek2009how, strait2014let, natarajan2020effects}. For instance, Hwang et al. \cite{HWANG2013459} found that the shape of a robot's individual components influences how its personality is perceived. 
Liberman-Pincu et al. \cite{Liberman2023} explored how the visual qualities of socially assistive robots influence users’ perceptions of their characteristics. Their findings suggest that user acceptance can be enhanced by selecting appropriate visual qualities and allowing minor customization. The authors note the importance of tailoring robot design to its intended role while accounting for individual user preferences.
Effectively describing and categorizing robot appearances can help researchers to identify meaningful links between design choices and human perceptions. Recently, Ringe et al. \cite{MetaMorph} introduced the \textit{MetaMorph} model, a comprehensive framework for systematically classifying and comparing visual features across various robot types. Unlike traditional broad categorization methods in HRI, which commonly focus on anthropomorphic and zoomorphic designs \cite{yanco2004classifying, baraka2020extended}, \textit{MetaMorph} provides a more inclusive and structured approach, enabling a broader analysis of robotic appearances.
Particularly in the kitchen setting, studies have shown that a robot's morphology can significantly impact user experiences by influencing perceptions, preferences, and engagement. For instance, a study by Zhu et al. \cite{zhu2020robot} suggests that increasing the anthropomorphism of a robotic chef enhances perceptions of warmth and competence, which in turn improves food quality prediction and mitigates negative attitudes toward robotic chefs replacing human chefs.
Although prior research highlights the importance of robot appearance on user interactions, little is known about how morphology affects user expectations and perceptions in collaborative cooking. This study adds to the body of research by examining how robots' visual design influences interaction in shared cooking tasks.

\subsection{Shared Spaces for Human-Robot Collaboration}
As human-robot collaboration becomes more common, designing shared spaces for seamless and safe interaction is increasingly important. 
Zhi and Lien et al. \cite{Zhi2025} introduced a computational approach to optimize kitchen layouts, utilizing a decentralized motion planner to efficiently coordinate human and robot movements, ultimately enhancing task performance.
However, one of the main concerns in human-robot collaboration is safety. As mentioned above, a kitchen setting presents significant potential risks, as a robot may need to handle sharp or heated objects.
Leusmann et al. \cite{Leusmann2023} examined how people perceive the danger level of kitchen objects when handled by themselves, another human, or a collaborative robot. Their results showed that objects were generally perceived as more dangerous when used by a collaborative robot. However, the authors noted three distinct danger categories emerging: sharp objects perceived as more dangerous when used by humans, neutral objects with no difference in perception, and fragile objects perceived as riskier when handled by a robot.
To measure the perceived danger of robots, Molan et al. \cite{Molan2025} developed and validated a 12-item Perceived Danger (PD) scale. The authors identified four key dimensions: affective states, physical vulnerability, ominousness, and cognitive readiness.
%
Our research contributes to the growing body of work on human-robot collaboration by examining how a robot's morphology influences its placement in a shared cooking environment and identifying the key factors users consider in making these decisions.

\section{STUDY DESIGN}
\label{sec:study_design}

Our study explores how morphological features of robots influences user expectations in collaborative tasks.
We focus on a mise en place scenario within a VR kitchen, where participants distribute tasks, arrange cooking items, and determine where to place a robot.
%
In the following, we detail the experimental setup, including the VR environment, the cooking tasks chosen, the varied robot designs, and the data collection methods.

\subsection{Study environment and materials}

\begin{figure*}[t]
    \centering
    \includegraphics[width=\linewidth]{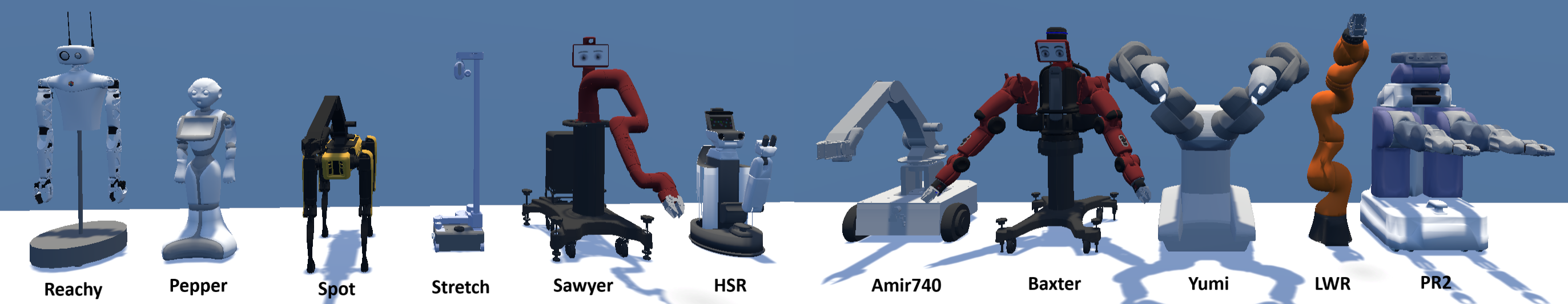}
    \caption{The eleven robots used in this study}
    \label{fig:used_robots}
\end{figure*}

\begin{figure}[t]
    \centering
    \includegraphics[width=\linewidth]{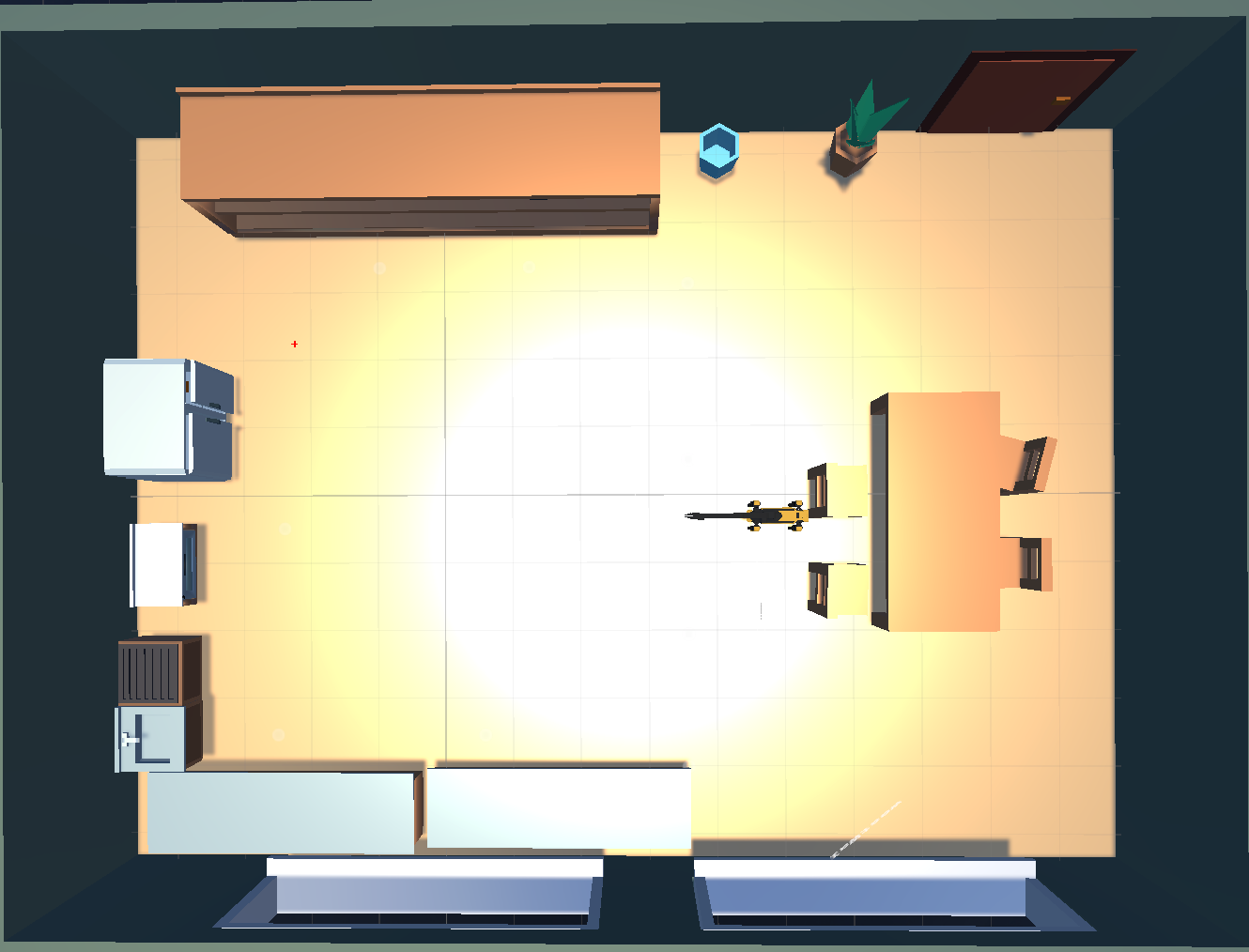}
    \caption{Top-down view of the kitchen environment}
    \label{fig:3D}
\end{figure}



\textbf{VR Setup:}  The virtual kitchen environment was built by synchronizing a Unity \cite{Unity} project with PyBullet \cite{coumans2021} and executed on a Meta Quest 3 head-mounted display (HMD). Figure \ref{fig:3D} shows a top-down 3D view of the virtual kitchen environment. All items -- ingredients or tools -- were initially located in the pantry at the left of or in the fridge at the bottom of the figure. The counters on the bottom have different heights, with the one near the sink being taller.

    
\textbf{Recipe and Tasks:} Two recipes -- a broccoli salad and almond cookies -- were selected from a published recipe corpus \cite{nevens2024benchmark} for their variety of tasks, tools, and ingredients. Both recipes include descriptions of ingredients, which we omit for space. The broccoli salad recipe and almond cookie recipe texts can also be found in the supplementary materials\footref{Suppl.mat.}.



\textbf{Robots:} Eleven robots, depicted in \Cref{fig:used_robots}, were chosen to represent a variety of sizes and morphological features, subject to the availability of suitable 3D models. Descriptions of the robots' morphology detailing each robot's features as classified by the \textit{MetaMorph} model \cite{MetaMorph} are available in the supplementary material\footref{Suppl.mat.}. Feature lists for all robots except for the Spot LWR robots were taken from the \textit{MetaMorph} dataset. The available \textit{MetaMorph} taxonomy was used to modify the Spot feature list from the dataset to reflect the added arm and gripper, as well as to create a list for the LWR robot. Using robots of various shapes was a conscious choice, even if some of them may not be usually thought of as kitchen robots.

\subsection{Participants}
\label{subsec:participants}

We recruited 23 participants aged 23--43 (11 male, 11 female, 1 undisclosed) via social media and word of mouth. Due to technical issues, recording data for participant 4 was lost so that our data set consisted of recordings of 22 participants in total. 
On a 7-point Likert scale, on average, they rated their robot experience slightly below medium ($\approx3.1$), VR experience slightly above medium ($\approx3.8$), and cooking experience above medium ($\approx5.3$).
Each participant was randomly assigned two different robots (one per recipe) so that no pair of robots was repeated. 



\subsection{Procedure}
\begin{figure}[tb!]
    \centering
    \begin{tikzpicture}[node distance=1.2cm]
    \tikzstyle{arrow} = [thick, ->, >=stealth]
    \tikzstyle{darrow} = [thick, dotted, ->, >=stealth]
    \tikzstyle{dashedprocess} = [process, dashed, fill=white]
    \tikzstyle{surround} = [process, fill=white]

    \begin{scope}[on background layer]
        \node (surroundbox) [surround, text depth=3.2cm, minimum height=4cm, inner xsep=0pt] at (0,-1) {\parbox{8cm}{\centering \textbf{Onboarding}}};
    \end{scope}
    
    \node (step1) [process] at (0,0) {\parbox{6cm}{\centering \textbf{Informed Consent}}};
    
    \node (step2) [dashedprocess, below of=step1] {\parbox{6cm}{\centering \textbf{VR kitchen tutorial}}};
    
    \node (step3) [process, below of=step2] {\parbox{6cm}{\centering \textbf{Demographics Questionnaire}}};
    
    \node (step4) [dashedprocess, below of=step3] {\parbox{8cm}{\centering \textbf{First recipe task}l}};
    
    \node (step5) [process, below of=step4] {\parbox{8cm}{\centering\textbf{First post-task interview}}};
    
    \node (step6) [dashedprocess, below of=step5] {\parbox{8cm}{ \centering \textbf{Second recipe task}}};
    
    \node (step7) [process, below of=step6] {\parbox{8cm}{ \centering \textbf{Second post-task interview}}};

    \node (step8) [process, below of=step7] {\parbox{8cm}{ \centering \textbf{Post-experiment interview}}};

    \draw [arrow] (step1) -- (step2);
    \draw [arrow] (step2) -- (step3);
    \draw [arrow] (step3) -- (step4);
    \draw [arrow] (step4) -- (step5);
    \draw [arrow] (step5) -- (step6);
    \draw [arrow] (step6) -- (step7);
    \draw [arrow] (step7) -- (step8);
    
    \end{tikzpicture}
    \caption{Study procedure steps. Dotted lines indicate steps conducted in VR}
    \label{fig:study_procedure}
\end{figure}
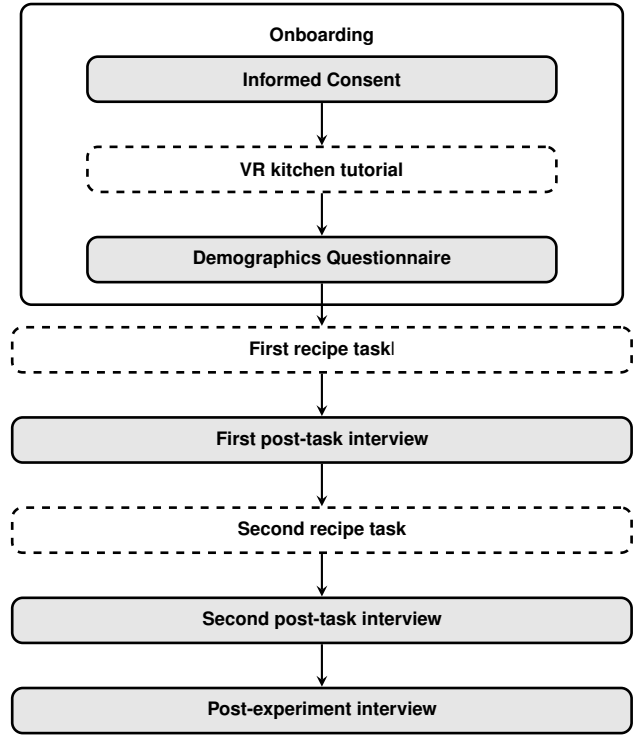

The study was conducted in a controlled laboratory environment.
TAPs and interviews were recorded via a clip-on microphone; the participants VR perspective were recorded via screen capture.
Two researchers supervised each session to ensure participants’ physical safety while immersed in VR, and to provide prompts or guidance when necessary.
The procedure consisted six phases that are visualized in figure \ref{fig:study_procedure} and further detailed in the following text.

\textbf{1.\ Onboarding phase:}
Participants first received information about the task and provided informed consent. They were told that they are to arrange the virtual kitchen as if cooking will take place afterwards, in collaboration with a robot. This robot would not move on its own during the study, but would be transported by the participant when they wish to find a location for the robot. The participants were told to act as if the robot is capable to perform any step of the recipe. 

Subsequently, participants completed a tutorial in the VR kitchen, familiarizing them with object interaction and movement to help standardizing the experience.
After this initial exposure, participants took a short break to reduce potential VR fatigue and filled out a demographics questionnaire (age, gender, prior experience with VR, robots, and cooking) as well as the General Attitude Towards Robots (GAToRS) questionnaire \cite{Koverola2022GeneralAT} to capture any biases.


\textbf{2.\ First Recipe Scenario:}
Participants re-entered the VR kitchen, where they were introduced to the broccoli salad recipe and their first robot.
They were instructed to position it and all necessary cooking items plus the robot according to their preferences to collaboratively prepare the recipe.
The participants were asked to narrate their thought process, particularly their rationale for item placement and task delegation to the robot versus doing a task autonomously.
After participants finalized their arrangement, they exited the VR environment.



\textbf{3.\ First Interview}
The participants took part in a brief semi-structured interview specific to this first recipe.
They were asked about their impressions of the kitchen layout, the rationale behind object and robot placement, their satisfaction with the setup, and potential changes they would suggest.
Participants were then offered an optional extended break to minimize VR-induced discomfort.


\textbf{4.\ Second Recipe Scenario}
Participants returned to the VR environment to repeat the mise en place scenario with a different recipe (almond cookies) and their second robot. Again, participants were instructed to position baking items and the robot according to preference and narrate their thought processes whilst doing so as well as specifying which tasks they planned on assigning to the robot and which they wanted to perform themselves.

\textbf{5.\ Second Interview}

After exiting the VR environment, another post-recipe questionnaire specific to this second recipe was again conducted to explore whether participants made different choices based on the second recipe or the new robot.


\textbf{6.\ Final Interview}

An additional structured post-experiment interview was conducted during which participants were asked if and how the robot's appearance and size ifluenced its placement and if they considered safety - the robot's as well as their own - when placing it. Additionally they were asked if they would prefer a stationary or mobile robot and to elaborate on the reasoning behind this preference. Lastly participants were encouraged to express any other comments.

\subsection{Annotation Scheme for Think-Alouds}
\label{subsec:annotation}

To address the questions of how the robots' morphology might have shaped the participants' decision-making regarding placements and task delegation, TAPs and Q\&A recordings were first transcribed and subjected to the following main annotation categories: 

\begin{enumerate}[label=\alph*.]
    \item Factors affecting placement of agents (robot or human). These included height of the agent, obstruction concerns, safety concerns, use of various robot features to make a decision, use of an emotional reaction to make a decision, willingness to share proximity with the robot.
    \item Factors affecting placement of items. These included the use of strategies such as an initial collection of items, grouping items by task, ordering by use order in a recipe, creation of gaps for future tasks or actions, and whether the participant perceived themselves unable to properly control the VR.
    \item Task delegation, which consisted of a list of who was assigned to the recipe subtasks. Options were robot, human, both, explicitly indifferent, or not specified.
\end{enumerate}

All variables were annotated in a binary-fashion. In addition, we annotated with free-form summaries of reasons provided by participants for task assignments, if any.

Two annotators have looked independently at the think-alouds from participants 1 to 8 (excluding 4). Intercoder agreement (Cohen's Kappa) scored  \( \kappa = 0.62 \) for category agent placement, \( \kappa = 0.69 \) for category item placement and \( \kappa = 0.85 \) for category task assignment. This was done to get an indication that the annotation categories can be reliably duplicated by new annotators. After this, one annotator continued annotating all other participants.

\section{RESULTS}
\label{sec:results}

\subsection{Quantitative Analysis of Task Assignments}

\begin{figure*}[tb]
  \centering

  \begin{subfigure}{0.31\textwidth}
    \centering
    \begin{tikzpicture}
      \begin{axis}[
        width=\textwidth+0.65cm+10pt,
        height=0.8\textwidth,
        ybar,
        bar width=5pt,
        axis x line=bottom,
        axis y line=left,
        trim axis left,
        trim axis right,
        clip=false,
        enlarge x limits={abs=10pt},
        inner xsep=0pt,
        inner ysep=0pt,
        yticklabel style={text width=2em, align=right},
        ylabel={Avg.\ Task Assignments},
        ylabel style={font=\scriptsize, yshift=-10pt, xshift=-6pt},
        symbolic x coords={Stretch,PR2,Baxter,Spot,Sawyer,LWR,Pepper,Reachy,Yumi,HSR,Amir},
        xtick=data,
        x tick label style={rotate=45,anchor=east,font=\tiny},
        ymin=0, ymax=5,
        enlarge y limits={lower,value=0.02},
          grid=both,
  major grid style={line width=0.3pt,draw=gray!40},
  minor grid style={line width=0.1pt,draw=gray!20},
      ]
        \addplot+[] coordinates {
          (Stretch,4.5) (PR2,4.25) (Baxter,4.0) (Spot,3.8) (Sawyer,2.75)
          (LWR,2.5) (Pepper,2.4) (Reachy,2.33) (Yumi,2.33) (HSR,2.0) (Amir,1.5)
        };
      \end{axis}
    \end{tikzpicture}
    \caption{Average number of tasks entrusted to each robot per participant. 
    }
    \label{fig:tasks}
  \end{subfigure}\hfill
  \begin{subfigure}{0.31\textwidth}
    \centering
    \begin{tikzpicture}
      \begin{axis}[
        width=\textwidth+0.65cm+2pt,
        height=0.8\textwidth,
        ybar,
        bar width=5pt,
        axis x line=bottom,
        axis y line=left,
        trim axis left,
        trim axis right,
          ymin=-0.1, 
        clip=false,
        enlarge x limits={abs=10pt},
        inner xsep=0pt,
        inner ysep=0pt,
        yticklabel style={text width=2em, align=right},
        ylabel={Avg.\ Envisioned Collab.},
        ylabel style={font=\scriptsize, yshift=-2pt, xshift=-6pt},
        symbolic x coords={Stretch,PR2,Baxter,Spot,Sawyer,LWR,Pepper,Reachy,Yumi,HSR,Amir},
        xtick=data,
        x tick label style={rotate=45,anchor=east,font=\tiny},
        ymin=0, ymax=1.5,
        enlarge y limits={lower,value=0.02},
          grid=both,
  major grid style={line width=0.3pt,draw=gray!40},
  minor grid style={line width=0.1pt,draw=gray!20},
      ]
        \addplot+[] coordinates {
          (Stretch,0.25) (PR2,0.5) (Baxter,0.25) (Spot,1.2) (Sawyer,0.5)
          (LWR,0.25) (Pepper,1.2) (Reachy,0.66) (Yumi,0.33) (HSR,0.0) (Amir,0.5)
        };
      \end{axis}
    \end{tikzpicture}
    \caption{Average number of cooperative tasks envisioned per participant. 
    }
    \label{fig:collabs}
  \end{subfigure}\hfill
  \begin{subfigure}{0.31\textwidth}
    \centering
    \begin{tikzpicture}
      \begin{axis}[
        width=\textwidth+0.65cm+2pt,
        height=0.8\textwidth,
        ybar,
        bar width=5pt,
        axis x line=bottom,
        axis y line=left,
        trim axis left,
        trim axis right,
          ymin=-0.1, 
        clip=false,
        enlarge x limits={abs=10pt},
        inner xsep=0pt,
        inner ysep=0pt,
        yticklabel style={text width=2em, align=right},
        ylabel={Collaborator Ratio},
        ylabel style={font=\scriptsize, yshift=-2pt, xshift=-6pt},
        symbolic x coords={Stretch,PR2,Baxter,Spot,Sawyer,LWR,Pepper,Reachy,Yumi,HSR,Amir},
        xtick=data,
        x tick label style={rotate=45,anchor=east,font=\tiny},
        ymin=0, ymax=1.0,
        enlarge y limits={lower,value=0.02},
          grid=both,
  major grid style={line width=0.3pt,draw=gray!40},
  minor grid style={line width=0.1pt,draw=gray!20},
      ]
        \addplot+[] coordinates {
          (Stretch,0.25) (PR2,0.25) (Baxter,0.25) (Spot,0.8) (Sawyer,0.5)
          (LWR,0.25) (Pepper,0.6) (Reachy,0.33) (Yumi,0.33) (HSR,0.0) (Amir,0.5)
        };
      \end{axis}
    \end{tikzpicture}
    \caption{Ratio of participants trusting the robot to collaborate on at least 1 task (max = 1)}
    \label{fig:ratio}
  \end{subfigure}

  \caption{Robot engagement metrics normalized by number of interacting participants, sorted by avg.\ task assignments (\ref{fig:tasks}).}
  \label{fig:threecol}
\end{figure*}
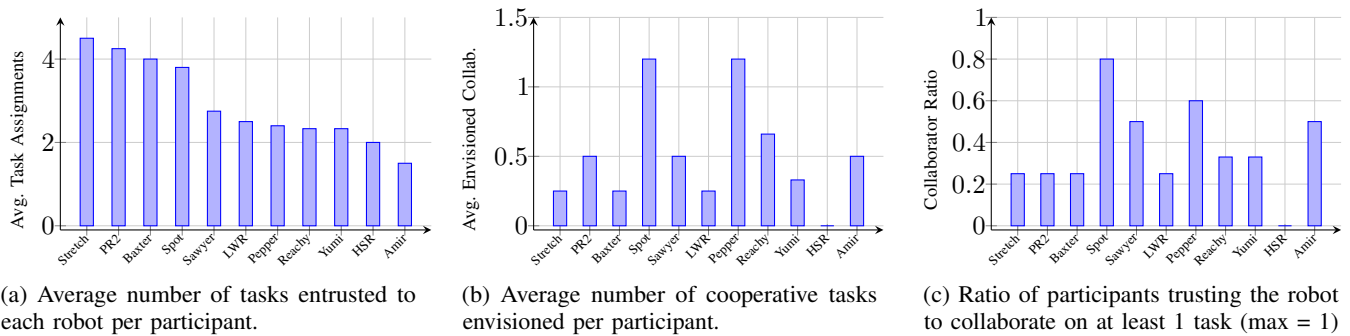

\Cref{fig:tasks} shows the average frequency with which each robot was assigned a task -- either by itself, in collaboration with the participant or because the participant was indifferent about who is doing a specific task -- relative to the number of participants who encountered the robot. Figure \ref{fig:collabs}
shows specifically how often participants, on average, envisioned collaborating with a robot on a task. In this context, collaboration was not defined in detail and could refer either to working simultaneously in close proximity or to taking turns at the same task. Lastly, \ref{fig:ratio} illustrates the ratio of participants who trusted the robot to collaborate on at least one task.

While the amount of data is too small to extract statistically robust conclusions, the diagrams in Figure \ref{fig:threecol} are suggestive of a distinction being made by the human participants between perceived competence and openness to collaboration. Thus, Stretch, PR2, and Baxter were allocated a good portion of the tasks but comparatively few participants wanted to share one of those tasks with them. Instead, Pepper and Spot were preferred by the participants when it came to sharing tasks. We hypothesize that it is their silhouette that is at least partially mimicking living beings -- the \textit{MetaMorph} model describes the silhouette for the Pepper as ``Humanoid Hybrid'' and ``Zoomorphic'' for Spot -- which may have contributed to this preference outcome.

\subsection{Qualitative Analysis of Task Assignments}

\paragraph{Perception of competence} Participants were told to assume the robots could do anything the recipe required in an effort to isolate their decisions from any experience they may have had with the current limitations of robots. This seems to have been interpreted as intended, i.e. the robots would be controlled competently enough to achieve anything physically possible with their bodyframes -- and in hindsight, we should have formulated our instruction in this way.

It is, however, interesting to note the manner in which the participants deviated from our stated instruction in more detail. As recorded in our think-alouds and interviews, participants tended to explicitly not consider robots suitable for tasks/placements when the robot was too small, or when its grippers appeared too simple and/or rigid to enable the fine motoric skills required for e.g. shaping dough. For example, some of the participants who saw the HSR explicitly judged it as too small to perform various tasks, which resulted in it being assigned less to do.

However, with a couple of exceptions, participants were not as concerned with the layout and abilities of a robot's cameras, and they would give tasks even to robots with no clear indication of visual sensors. Spot, while popular to share tasks with, has no clearly visible cameras that could also swivel or otherwise get a wider view of its surroundings. This suggests that humans are more ready to assume hidden sensory competences, but will assume that capabilities for action are matched closer by visual aspect.



\paragraph{Strategies for sharing space} We consider here how the participants envisioned sharing the space of the kitchen, not necessarily in close proximity to the robot, however. As the lack of total collaborations depicted in \Cref{fig:collabs} suggests, participants rarely envisioned sharing proximity with the robot and often took measures to avoid it. This tendency to avoid close interaction is further supported by our additional annotations of the TAPs (not shown here due to space constraints), which indicate that out of the 22 participants, only 7 reported sharing their workspace in close proximity with the robot during the first recipe, and only 3 during the second.  They sometimes organized the workspace in distinct areas with the human workspace out of the robot's reach. One participant said they would have liked to place the robot across the counter from them, to have a physical barrier between the robot and themselves. A few expressed an expectation that the robot would back off after finishing a task. And, despite item transport being often given to the robot, a few participants expressed a preference for a robot that is static at least when performing a task.


Safety was sometimes a reason behind such strategies; however, participants also often reported not regarding safety as a concern. Instead, they considered the possibility of the robot getting in their way, or vice-versa, or simply having a handle on where the robot will be at any given moment. We suspect the underlying issue is whether the robot would be a welcome presence in the space or not. Even if on the whole, larger robots may be perceived as more capable, their sheer bulk makes them appear as an annoyance to route around. We note that the three robots with the highest number of collaborative assignments from \Cref{fig:collabs} are more gracile -- i.e., less bulky -- than most of the others, which may suggest they are physically less capable to be an obstruction -- and may also suggest to the human that such robots are more capable to steer out of the way by themselves. Note, gracile is not necessarily small (e.g. Spot was fairly large in our environment), but rather slender. Emotional reactions at the shapes of more technomorphic robots such as Amir740, LWR, and Yumi also dissuaded some participants from using them in close proximity or at all. 

\section{DISCUSSION, LIMITATIONS, AND FUTURE WORK}
\label{sec:discussion}

Our study here has been an exploratory one, looking for what kind of factors about a robot's morphology may be decisive for whether, and how, a human would choose to share space and perform tasks together with that robot. Thus, we do not present findings but rather hypotheses suggested by our observations, which we intend to follow up on in larger, subsequent studies. Our analysis of the data provided by the participants, described in Section~\ref{sec:results}, suggested a difference between human perceptions of competence and openness to collaboration, as well as a difference in how sensoric and motoric capabilities are perceived. For the competence/collaboration topic, robot size appears associated to perceptions of competence. It is less associated to openness for collaboration, as the more popular robots include both relatively large and relatively small ones. Thus, we had to look at other features of the popular robots to identify what they may have in common, and so formulated our hypotheses:

\begin{itemize}
    \item \textbf{H1} A more biomorphic morphology will encourage humans to share tasks with a robot.
    \item \textbf{H2} Humans tend to assume sensoric capabilities based on the action capabilities suggested by a robot's body. E.g., if it has a hand it can probably feel by touch, if it has an arm it can swing it and can probably see where to swing it, etc.
    \item \textbf{H3} Gracile robots will be avoided less than more physically imposing ones.
\end{itemize}

We have conducted our study in VR, with robots that are static props and instructed the participants to imagine how they would collaborate with these robots. While this allowed us to include several robot shapes and focus on robot morphology as opposed to observed behavior, it also brings some drawbacks. The VR environment was not always easy to control by the participants, resulting in arrangements of items that were less precise than the participants intended. 
Static robot models were used to isolate the impact of morphology without the influence of movement or behavior. This allowed us to compare various robot forms based solely on appearance. However, this also required participants to imagine the robots' movement and capabilities, which may have introduced interpretive variability. Some participants also faced minor challenges with object placement in VR. Despite these limitations, using static robots was appropriate for this exploratory study and helped generate initial hypotheses for future work.
Keeping in mind both the arranging of items and imagining what the robot may be able to do, is a complex cognitive task that may skew results. Several participants expressed a desire for the robot to help them set up the environment. Thus, for future studies, we are considering to not only increase the number of participants, but also what improvements we could make to the VR interface, as well as whether to allow the VR robots to perform some tasks in a wizard of Oz setup where a hidden researcher teleoperates the robot avatar.

\bibliographystyle{IEEEtran}  

\bibliography{references}     

\end{document}